\pdfoutput=1

\documentclass[11pt]{article}
\usepackage[final]{acl}
\usepackage[T1]{fontenc}
\usepackage[utf8]{inputenc}
\usepackage{times}
\usepackage{latexsym}
\usepackage{microtype}
\usepackage{inconsolata}
\usepackage{amsmath}
\usepackage{amssymb}
\usepackage{graphicx}
\usepackage{booktabs}
\usepackage{tcolorbox}
\usepackage{multirow}
\usepackage{xcolor}
\usepackage{url}
\PassOptionsToPackage{unicode=false}{hyperref}
\usepackage{hyperref}
\usepackage{tabularx}
\usepackage[arabic,farsi,english]{babel}
\usepackage[table]{xcolor}
\usepackage[table]{xcolor}
\usepackage{tabularx}
\usepackage{booktabs}
\usepackage{array}
\newcolumntype{L}[1]{>{\raggedright\arraybackslash}p{#1}}
\definecolor{lightgray}{gray}{0.95}
\definecolor{softred}{rgb}{1,0.9,0.9}
\definecolor{softgreen}{rgb}{0.9,1,0.9}

\definecolor{darkgreen}{rgb}{0.0, 0.5, 0.0}
\definecolor{darkred}{rgb}{0.5, 0.0, 0.0}

\title{
\bfseries \textsc{FFE-Hallu}: Hallucinations in Fixed Figurative Expressions\\
Benchmark of Idioms and Proverbs in the Persian Language
}

\date{}

\author{
Faezeh Hosseini$^{*1}$ \quad
Mohammadali Yousefzadeh$^{*1}$ \quad
Yadollah Yaghoobzadeh$^{1,2}$ \\
$^{1}$Tehran Institute for Advanced Studies, Khatam University, Iran \\
$^{2}$School of Electrical and Computer Engineering,\\
College of Engineering, University of Tehran, Tehran, Iran \\
\texttt{f.hosseini401@khatam.ac.ir} \quad
\texttt{m.yousefzadeh@teias.institute} \quad
\texttt{y.yaghoobzadeh@ut.ac.ir} \quad \\
}

\begin{document}
\maketitle
\begin{abstract}
Figurative language, particularly fixed figurative expressions (FFEs) such as idioms and proverbs, poses persistent challenges for large language models (LLMs). Unlike literal phrases, FFEs are culturally grounded, largely non-compositional, and conventionally fixed, making them especially vulnerable to \emph{figurative hallucination}. We define figurative hallucination as the generation or endorsement of expressions that sound idiomatic and plausible but do not exist as authentic figurative expressions in the target language. 
We introduce \textsc{FFE-Hallu}, the first comprehensive benchmark for evaluating figurative hallucination in LLMs, with a focus on Persian—a linguistically rich yet underrepresented language. \textsc{FFE-Hallu} consists of 600 carefully curated instances spanning three complementary tasks: (i) FFE generation from meaning, (ii) detection of fabricated FFEs across four controlled construction categories, and (iii) FFE-to-FFE translation from English to Persian. 
Evaluating six state-of-the-art multilingual LLMs, we find systematic weaknesses in figurative competence and cultural grounding. While models such as GPT-4.1 demonstrate relatively strong performance in rejecting fabricated FFEs and retrieving authentic ones, most models struggle to reliably distinguish real expressions from high-quality fabrications and frequently hallucinate during cross-lingual translation. These findings reveal substantial gaps in current LLMs’ handling of figurative language and underscore the need for targeted benchmarks to assess and mitigate figurative hallucination.
\end{abstract}

\section{Introduction}
Large Language Models (LLMs) have achieved remarkable success across many NLP tasks, yet they continue to struggle with figurative language—expressions whose meanings cannot be derived compositionally from their constituent words. In this work, we focus on a specific subset of figurative language, which we term \emph{fixed figurative expressions} (FFEs), encompassing idioms and proverbs. FFEs are conventionally fixed, permitting only limited surface variation (e.g., tense or inflection), while preserving core lexical and structural elements that define their idiomatic identity \citep{nunberg1994idioms}. For instance, in the idiom ``kick the bucket,'' the lexical items ``kick'' and ``bucket'' are essential regardless of syntactic context. Similarly, in the proverb ``Actions speak louder than words,'' the key elements and their order are largely invariant. FFEs are therefore both highly conventionalized and largely non-compositional, as well as deeply culturally grounded, making them particularly challenging for LLMs to interpret or generate accurately \citep{kabra-etal-2023-multi,chakrabarty2022flute}.

In natural communication, FFEs enrich spoken and written language by conveying condensed meaning, evaluative stance, humor, and cultural resonance. Their meanings are often opaque and cannot be reliably recovered through literal or word-by-word translation \citep{de-luca-fornaciari-etal-2024-hard,mi-etal-2025-rolling,rezaeimanesh-etal-2025-large}. As a result, robust handling of FFEs is critical for applications such as machine translation, dialogue systems, and cross-cultural or educational technologies, all of which require not only linguistic fluency but also cultural competence.

While factual hallucination in LLMs has been widely studied \citep{ji2023survey}, hallucination in figurative language remains largely unexplored. We define \emph{figurative hallucination} as the generation or endorsement of non-existent fixed figurative expressions (FFEs) that sound idiomatic and plausible. Such hallucinations can appear convincingly authentic, propagate cultural misinformation, and undermine trust in model outputs. This risk is amplified by the illusory truth effect: familiar-sounding statements—especially those resembling established idiomatic templates—are more likely to be perceived as true, even by informed readers \citep{fazio2015knowledge}. Consequently, non-native and novice users are particularly vulnerable to fabricated FFEs. As reliance on LLM-generated text continues to grow, addressing figurative hallucination becomes increasingly important.

To address this gap, we introduce \textsc{FFE-Hallu} (Fixed Figurative Expressions Hallucination), a benchmark designed to evaluate figurative and cultural competence in LLMs, with a focus on hallucination behavior. \textsc{FFE-Hallu} comprises three complementary tasks: 
(i) \emph{FFE Generation from Meaning}, which tests whether a model can retrieve a valid idiom or proverb given a figurative definition; 
(ii) \emph{Fake FFE Detection}, which assesses the ability to distinguish authentic expressions from plausible fabrications; and 
(iii) \emph{FFE-to-FFE Translation}, which evaluates whether models can produce culturally and semantically appropriate idiomatic equivalents across languages, rather than resorting to literal or invented expressions.

The benchmark comprises 600 curated instances across three settings: 200 authentic idioms and proverbs, 200 carefully constructed fabrications simulating figurative hallucinations, and 200 bilingual English–Persian idioms with verified equivalents. Together, these tasks provide a systematic framework for probing how LLMs handle figurative meaning, cultural grounding, and hallucination. Because figurative hallucinations concern culturally grounded expressions rather than verifiable facts, it is unclear whether commonly used LLM-as-a-judge methods can reliably assess them. We therefore explicitly investigate the validity of automatic evaluation for FFEs in this work.

This paper makes the following contributions:
(i) We introduce \textsc{FFE-Hallu}, the first benchmark specifically designed to evaluate figurative hallucination in LLMs, focusing on fixed figurative expressions in Persian.
(ii) We propose three complementary tasks—FFE generation from meaning, fake FFE detection, and FFE-to-FFE translation—that probe distinct aspects of figurative competence and cultural grounding.
(iii) We evaluate six state-of-the-art multilingual LLMs, both open-weight and proprietary, and provide a detailed analysis of their vulnerabilities in figurative generation, detection, and cross-lingual transfer.
(iv) We assess the reliability of LLM-as-a-judge evaluation for figurative hallucination by comparing automatic judgments against expert human annotations.

\section{Related Work}

We organize this section into two main areas: (1) hallucination and reliability in language models, focusing on definitions and taxonomies relevant to our study, and (2) research on figurative language processing, particularly idiom and proverb comprehension. This organization situates \textsc{FFE-Hallu} at the intersection of these lines of work and highlights the novelty of systematically evaluating \emph{figurative hallucination}.

\subsection{Hallucination and Reliability in Language Models}

Hallucination—broadly defined as the generation of content not grounded in the input or the model’s training knowledge—has emerged as a major challenge in natural language generation \citep{ji2023survey}. Prior work commonly distinguishes between \emph{intrinsic} hallucinations, which contradict the input, and \emph{extrinsic} hallucinations, which introduce information unsupported by known facts \citep{rawte2023troubling}. Other studies differentiate between \emph{factual} and \emph{non-factual} hallucinations \citep{cao2022hallucinated}. Factual hallucinations refer to content that is not directly inferable from the source but is verifiable and correct according to external world knowledge, whereas non-factual hallucinations involve entities or claims that are neither supported by the source nor factually valid \citep{bendahman2025not}.

Our definition of hallucination in the context of fixed figurative expressions (FFEs) falls within the category of extrinsic, non-factual hallucination. Specifically, we consider cases in which models generate or endorse fabricated FFEs. Unlike factual hallucinations, such errors cannot be reliably identified through external fact checking and instead require cultural and linguistic grounding for accurate assessment.

\subsection{Figurative Language and Idiomaticity}

Idioms are a core component of figurative language, characterized by non-compositional meanings that pose challenges for both rule-based and data-driven systems. Recent work by \citet{de-luca-fornaciari-etal-2024-hard} shows that idiom detection remains “a hard nut to crack” even for conversational LLMs, which continue to struggle to distinguish literal from figurative usage. Similarly, \citet{mi-etal-2025-rolling} introduce \textsc{DICE}, a framework for evaluating idiom comprehension through contextual understanding, and demonstrate that even state-of-the-art models perform poorly when idiomatic expressions appear in context.

The multilingual dimension of idiomaticity has been explored in benchmarks such as \textsc{MABL} \citep{kabra-etal-2023-multi}, which evaluates figurative language understanding across eight languages, as well as in comparative studies of idioms and similes in LLMs \citep{khoshtab-etal-2025-comparative}.

For Persian, \citet{rezaeimanesh-etal-2025-large} introduce \textsc{PersianIdioms}, a parallel corpus of 2{,}200 idioms—700 with usage examples—and evaluate idiomatic translation using both LLM-based and hybrid systems, reporting strong performance from Claude~3.5~Sonnet. Recent work has also proposed metrics and self-checking mechanisms for assessing idiomatic metaphor competence \citep{gao2025crst}.

Our work complements this literature by shifting the focus from idiom understanding or translation accuracy to \emph{hallucination behavior} in figurative language. \textsc{FFE-Hallu} is the first benchmark designed to evaluate whether LLMs can distinguish authentic fixed figurative expressions from plausible fabrications and generate culturally valid FFEs from meaning, providing a targeted diagnostic of figurative and cultural grounding.

\section{\textsc{FFE-Hallu} Benchmark}
\label{sec:benchmark}

Hallucination in LLMs has been extensively studied in factual domains \citep{ji2023survey,rawte2023troubling}. In this work, we focus on a distinct and underexplored subtype: \emph{figurative hallucination}. 

This notion is distinct from \emph{semantic errors}, in which a model produces an existing FFE whose meaning does not align with the intended interpretation. In our benchmark, such cases are labeled as incorrect rather than hallucinated, and our annotation protocol explicitly enforces this distinction.

While LLMs are capable of creative linguistic behavior, our benchmark evaluates knowledge of authentic idioms and proverbs rather than creative invention. Accordingly, when a model produces a non-existent expression, we categorize it as a hallucination rather than creativity, as such outputs risk misleading users into treating fabricated expressions as culturally valid. 
Unlike factual hallucinations, which can often be verified through external sources, figurative hallucinations require deeper linguistic and cultural grounding to be reliably identified and evaluated.

The goal of \textsc{FFE-Hallu} is to systematically probe figurative hallucination in LLMs through three complementary tasks. Each task targets a distinct dimension of FFE competence—\textit{generation}, \textit{detection}, and \textit{translation}—while collectively capturing both the prevalence and characteristics of figurative hallucinations. The following sections describe these tasks in detail.

\subsection{Task 1: FFE Generation from Meaning}
\label{subsec:task1}

\paragraph{Motivation}
Most prior work on idiom and proverb processing has focused on detection or translation, rather than generation from meaning. Generating a valid fixed figurative expression (FFE) from its figurative definition constitutes a stronger test of a model’s internal idiomatic knowledge: it requires retrieving an established, culturally conventionalized expression rather than constructing a novel one. Conventional text generation benchmarks primarily assess factual recall or paraphrasing ability and therefore do not evaluate whether a model can produce semantically appropriate yet fixed figurative forms.

This task introduces a new diagnostic perspective by measuring whether a model can successfully retrieve authentic idioms or proverbs for a given figurative meaning, or instead hallucinates by producing non-existent expressions.

\paragraph{Task Definition}
In this task, models are given a figurative definition in Persian and asked to produce an idiom or proverb (i.e., an FFE) that conveys the same meaning. The central challenge is to generate an expression that is both semantically appropriate and culturally valid. To reduce potential data contamination and memorization effects, we evaluate models using two types of definitions: (i) human-authored figurative meanings sourced from the Abadis online dictionary \citep{rezaeimanesh-etal-2025-large}, and (ii) GPT-4o-generated paraphrases that are manually verified to ensure they introduce no additional information beyond the original meaning.

Model outputs arecategorized into three classes: (a) \textbf{correct}, where the generated expression is an authentic FFE that matches the intended meaning; (b) \textbf{incorrect}, where the output is a real FFE but semantically mismatched; and (c) \textbf{hallucinated}, where the model produces a literal phrase or fabricates a non-existent expression. Table~\ref{tab:idiom-failures} illustrates these two distinct failure modes in Task~1: generating an existing but semantically incorrect FFE, and hallucinating a non-existent FFE.

\begin{table*}[t]
\centering
\renewcommand{\arraystretch}{1.25}
\small
\begin{tabular}{p{0.18\linewidth} p{0.38\linewidth} p{0.38\linewidth}}
\toprule
\textbf{Failure Type} & \textbf{Real FFE but Incorrect} & \textbf{Hallucinated FFE} \\
\midrule
\textit{Input Meaning} & 
Regretting a wrong action and making 

a firm decision not to repeat it. &
Someone who considers themselves superior to others. \\
\midrule
\textit{Ground Truth Idiom} & 
\FR{پشت دست خود را داغ کردن}

\textit{Gloss:} ``To burn the back of one’s hand''

\textit{Meaning:} To strongly vow not to repeat

a mistake. &
\FR{تافته‌ی جدا بافته}

\textit{Gloss:} ``A special thread spun separately''

\textit{Meaning:} A person who sees themselves as

exceptional. \\
\midrule
\textit{Model Output} & 
\FR{توبه شکستن}

\textit{Gloss:} ``To break a vow''

\textit{Meaning:} Implies repetition of the mistake. &
\FR{موش توی انبار کاه}

\textit{Gloss:} ``A mouse in a haystack''

\textit{Meaning:} - \\
\bottomrule
\end{tabular}
\caption{Examples of model failures in Task 1: FFE Generation. Each example includes the Persian FFE, its English gloss (literal translation), and its figurative meaning.}
\label{tab:idiom-failures}
\end{table*}

\subsection{Task 2: Fake FFE Detection}
\label{subsec:task2}

\paragraph{Motivation}
Inspired by the \textit{NonExistentRefusal} task in HalluLens~\citep{bang2023hallulens}, which examines whether LLMs hallucinate information about non-existent entities, this task targets a fundamentally different phenomenon: \textit{figurative hallucination}. Rather than assessing factual knowledge, we test whether models can recognize \textit{synthetic idioms and proverbs}—expressions that are structurally and semantically plausible but do not exist in the target language. This extends hallucination evaluation beyond factual correctness toward probing figurative and cultural understanding. In contrast to factual verification tasks, where correctness can be checked against external sources, fake FFE detection requires models to distinguish authentic idiomatic expressions from fabricated yet convincing ones, demonstrating true competence in idiomatic form, meaning, and usage.

\paragraph{Task Definition}
In this task, models are prompted to assess a given FFE and determine whether it is authentic or fabricated. Each model is presented with artificially constructed FFEs designed to closely mimic the syntax, tone, and metaphorical style of real idioms and proverbs. To evaluate the reliability of rejection behavior, we employ two complementary prompt formats: a \textit{Fake-detection} prompt (“\textit{Is this FFE fake in Persian?}”), where the correct answer is \textit{Yes}, and a \textit{Real-confirmation} prompt (“\textit{Is this FFE real in Persian?}”), where the correct answer is \textit{No}. Each fabricated FFE is tested under both prompt types to examine consistency across framing conditions. We define the hallucination rate as the proportion of cases in which the model incorrectly accepts a fake FFE as real. Responses are labeled as (i) \textbf{correct} if the model detects and rejects the fabricated expression, and (ii) \textbf{hallucinated} if it endorses a fake expression as genuine.

The fabricated FFEs used in this task were deliberately constructed to resemble authentic Persian idioms and proverbs in form and metaphorical structure, while remaining culturally invalid.
They were generated using controlled strategies grounded in linguistic research, including semantic inversion, lexical perturbation, and structural mimicry. A detailed description of the dataset construction process and the four fabrication categories is provided in Section~\ref{subsec:dataset}.

\begin{table*}[t]
\centering
% \small
\begin{tabular}{lccc}
\toprule
\textbf{Task} & \textbf{Source} & \textbf{\# Items} & \textbf{Goal} \\
\midrule
FFE Generation & Authentic Persian idioms/proverbs & 200 & Cultural recall from meanings \\
Fake FFE Detection & Fabricated idioms (4 categories) & 200 & Rejection of hallucinations \\
FFE Translation & English $\rightarrow$ Persian pairs & 200 & Cross-lingual hallucination test \\
\midrule
\textbf{Total} & --- & \textbf{600} & --- \\
\bottomrule
\end{tabular}
\caption{Dataset composition of \textsc{FFE-Hallu}, grouped by task.}
\label{tab:dataset-overview}
\end{table*}

\subsection{Task 3: FFE Translation (English $\rightarrow$ Persian)}
\label{subsec:task3}

\paragraph{Motivation}
Cross-lingual idiom translation poses a significant challenge for LLMs, as the correct output cannot be achieved through literal, word-for-word translation. Instead, the model must retrieve or generate an equivalent expression that preserves the same figurative meaning while fitting the target culture. Recent work has begun to tackle this problem; for example, \citet{donthi2025idiomatic} improve idiomatic translation by linking idioms across languages. However, such studies primarily focus on identifying correct equivalents and overlook the issue of \textit{figurative hallucination}—cases where a model fabricates invalid idioms or proverbs. 

This task reframes idiom translation as a test of both cultural and figurative competence, assessing whether models can transfer idiomatic meaning across languages without resorting to literal or invented expressions. In doing so, it bridges figurative reasoning and hallucination analysis, highlighting the gap between semantic accuracy and cultural authenticity.

\paragraph{Task Definition}
In this task, models are given an English idiom or proverb that has at least one verified Persian equivalent and are asked to produce a culturally equivalent FFE in Persian.
Model outputs are categorized as follows: (i) \textbf{correct}, if the model produces a valid Persian FFE whose figurative meaning accurately matches that of the source expression; (ii) \textbf{incorrect}, if the model outputs an existing Persian FFE whose figurative meaning does not align with the source; and (iii) \textbf{hallucinated}, if the model generates a literal, word-for-word translation or an invented expression that does not exist as a Persian FFE. Since models are explicitly instructed to produce an equivalent figurative expression rather than a literal translation, word-for-word outputs are treated as hallucinations. Such outputs may appear fluent and plausible, yet they introduce non-existent FFEs into the language and risk misleading non-native speakers, thereby undermining trust in model-generated content.
\subsection{Dataset Construction}
\label{subsec:dataset}
To systematically study figurative hallucination, we introduce \textsc{FFE-Hallu}, a 600-item benchmark combining authentic and adversarially constructed cases across the three tasks described in Sections~\ref{subsec:task1}–\ref{subsec:task3}. For Task~1, 200 real FFEs were collected from online Persian dictionaries with their figurative meanings \citep{rezaeimanesh-etal-2025-large}.  

For Task~2, 200 fabricated FFEs were constructed to resemble authentic Persian idioms and proverbs while remaining culturally invalid.
These expressions fall into four controlled categories, designed to probe different sources of figurative hallucination. The first two categories preserve the surface template of real FFEs and are created manually: \textit{(1)Word Perturbation}; modifies an existing FFE by replacing a key lexical item (e.g., noun or verb) with a semantically related or phonologically similar word, yielding a fluent but non-existent expression. \textit{(2)Semantic Inversion}; retains the original idiomatic structure while reversing or distorting the figurative meaning, producing expressions that appear idiomatic but encode an incompatible concept.

The remaining two categories are \emph{fully synthetic} and do not originate from any real FFE and  are created automatically using GPT-4o: \textit{(3)Structural/Syntactic Mimicry}; generates expressions that follow common proverb-level constructions in Persian—such as parallelism, contrast, or cause–effect framing—without borrowing lexical material from existing idioms. \textit{(4)Cultural/Historical Fabrication} constructs FFE-like expressions using culturally salient elements (e.g., rituals, historical figures, occupations, symbolic objects, or social occasions) that frequently appear in Persian proverbs.

These elements and structural patterns are informed by linguistic analyses of Persian idiom and proverb anatomy \citep{zolfaghari2012persian}. GPT-4o was used only to generate candidate expressions following these templates, and all outputs were manually reviewed and refined by expert annotators to ensure plausibility and non-existence. Further details and examples for each category are provided in Appendix~\ref{appendix:dataset-creation}.

% All fabricated FFEs were verified to be absent from major Persian dictionaries and corpora.
% Candidate expressions were generated using controlled templates and subsequently reviewed and refined by expert annotators to ensure plausibility, grammaticality, and non-existence.

% For Task~2, 200 fabricated FFEs were created to mimic authentic idioms and proverbs while remaining culturally invalid (absent from all major dictionaries and corpora). They fall into four categories: (1) \textit{Semantic Inversion}, (2) \textit{Word Perturbation}, (3) \textit{Structural/Syntactic Mimicry}, and (4) \textit{Cultural/Historical Fabrication}. The first two were derived from real FFEs by reversing meaning or replacing a key word. The latter two were fully synthetic, designed using structural and cultural patterns observed in real Persian proverbs \citep{zolfaghari2012persian}. GPT-4o was used only to generate candidate expressions following these templates, and all outputs were manually reviewed and refined by expert annotators to ensure plausibility and non-existence.

For Task~3, 200 bilingual English–Persian idiom pairs with verified figurative equivalence were gathered from online resources and validated by native Persian speakers fluent in English.

The complete dataset is publicly available.\footnote{\url{https://github.com/hfaezeh77/FFE-Hallu-Benchmark}}

\subsection{Annotation Protocol}
All data entries and model outputs were reviewed by two native Persian speakers with expertise in figurative language. For authentic FFEs in Task~1, annotators verified the accuracy of definitions and usage examples. For fabricated FFEs in Task~2, they confirmed that none appear in dictionaries or natural usage. Fake FFEs generated by GPT-4o were additionally checked to ensure they closely resemble real idioms and proverbs in form, style, and cultural references, while remaining entirely invalid in Persian. For bilingual pairs in Task~3, annotators verified correct figurative equivalence, proposed improved translations where needed, and removed incorrect pairs. All disagreements were resolved through discussion with a third annotator.

\renewcommand{\arraystretch}{1.3}
\begin{table*}[t]
\centering
\resizebox{\textwidth}{!}{%
\begin{tabular}{lcccccc}
\toprule
\textbf{Model} & 
\multicolumn{3}{c}{\textbf{Human-Written Meaning}} & 
\multicolumn{3}{c}{\textbf{LLM-Paraphrased Meaning}} \\
\cmidrule(lr){2-4} \cmidrule(lr){5-7}
 & \textbf{Correct(\%)~$\uparrow$} & \textbf{Incorrect(\%)} & \textbf{Hallucination(\%)~$\downarrow$} & \textbf{Correct(\%)~$\uparrow$} & \textbf{Incorrect(\%)} & \textbf{Hallucination(\%)~$\downarrow$} \\

\midrule
GPT-4.1            & \textbf{\textcolor{darkgreen}{60.5}} & 19.0 & 20.5 & 59.5 & 23.5 & 17.0  \\
Claude 3.7 Sonnet  & 50.0 & 35.0 & 15.0 & 53.0 & 33.0 & \textbf{\textcolor{darkgreen}{14.0}} \\
Gemma3-27B        & 22.5 & 38.0 & 39.5 & 30.5 & 35.5 & 34.0  \\
DeepSeek-R1        & 48.5 & 24.5 & 27.0 & 47.0 & 27.5 & 25.0   \\
DeepSeek-V3        & 17.5 & 40.5 & 42.0 & 18 & 24.5 & 47.5 \\
Qwen3-235B          & \textbf{\textcolor{red}{7.0}} & 23.5 & \textbf{\textcolor{red}{69.5}} & 10.0 & 21.0 & 69.0 \\
\bottomrule
\end{tabular}
}
\caption{Performance of LLMs on Task 1 described in ~\ref{subsec:task1}: FFE Generation, from two input types: Human-Written Meanings and LLM-Paraphrased Meanings. Metrics include correct idiom generation percentage, incorrect generation, and hallucination rate. Higher Correct(\%) and lower Hallucination(\%) are desirable.}
\label{tab:idiom-gen-results}
\end{table*}

\section{Experimental Setup and Results}
\label{sec:setup}

\paragraph{Models}
We evaluate six large language models selected for their multilingual coverage and demonstrated strength on reasoning and generation tasks. We prioritized models with strong overall performance to ensure that observed failures reflect figurative competence rather than general model weakness. The evaluated models include GPT-4.1 \citep{openai_gpt4.1_2025,openai2023gpt4}, Claude~3.7 Sonnet \citep{anthropic_claude3.7_2025}, Gemma3-27B \citep{gemma3_arxiv_2025}, DeepSeek-V3 and DeepSeek-R1 \citep{deepseek_v3_r1_2025}, and Qwen3-235B \citep{qwen3_arxiv_2025}. Together, these models span both proprietary API-based systems and open-weight releases, as well as a wide range of parameter scales.
%The parameter counts for each model, as reported by their official documentation, are: GPT-4.1 (undisclosed), Claude 3.7 Sonnet (undisclosed), Gemma3-27B (27B), DeepSeek-V3 (236B), DeepSeek-R1 (67B), Qwen3-235B (235B).

\paragraph{Computational Resources}
All LLM outputs were obtained via official API endpoints. Across all experiments, total API usage amounted to approximately \$80 USD. All evaluations were performed on the cloud using OpenAI and Anthropic API services; no additional GPU or local compute resources were required.

\paragraph{Prompts}
All models are queried with consistent, task-specific prompt templates to ensure comparability across tasks. Prompts are designed to be concise, neutral in style, and free of additional examples beyond what is necessary for task clarification. For Task 1: FFE generation, the input is always a concise figurative meaning in Persian; for Task 2: Fake FFE Detection, a fake FFE; and for Task 3: FFE to FFE translation, an English idiom or proverb that has at least a known Persian equivalent. Full prompt templates for each task are provided in Appendix~\ref{appendix:Prompt-Templates}.

\paragraph{Evaluation Protocol}
We employ both human and automatic evaluation across the three tasks. 

\textit{Human Annotation (Tasks 1 and 3);}
For the FFE generation and FFE-to-FFE translation tasks, all model outputs were labeled using the same three-category scheme defined in Sections~\ref{subsec:task1} and~\ref{subsec:task3}. All model outputs were labeled by two native Persian annotators with disagreements resolved by a third annotator. Inter-annotator agreement (IAA) was calculated to assess consistency; detailed scores are reported in Appendix~\ref{appendix:annotation}.

\textit{Automatic Ground-Truth Evaluation (Task 2);}
For the Fake FFE Detection task, evaluation was automatic, based on model responses of ``YES'' or ``NO''. A response is marked \textit{correct} if the model correctly identifies a fake FFE as fake, otherwise \textit{hallucinated}.

\textit{LLM-as-a-Judge Evaluation (Tasks 1 and 3);}  
To explore whether automatic evaluation can approximate human annotation quality, we employed GPT-4.1 and Gemini~2.5~Pro as LLM-as-a-judge. Full evaluation details and results are provided in Section~\ref{sec:auto-eval}.

\begin{figure*}[ht]
    \centering
    \includegraphics[width=1\linewidth]{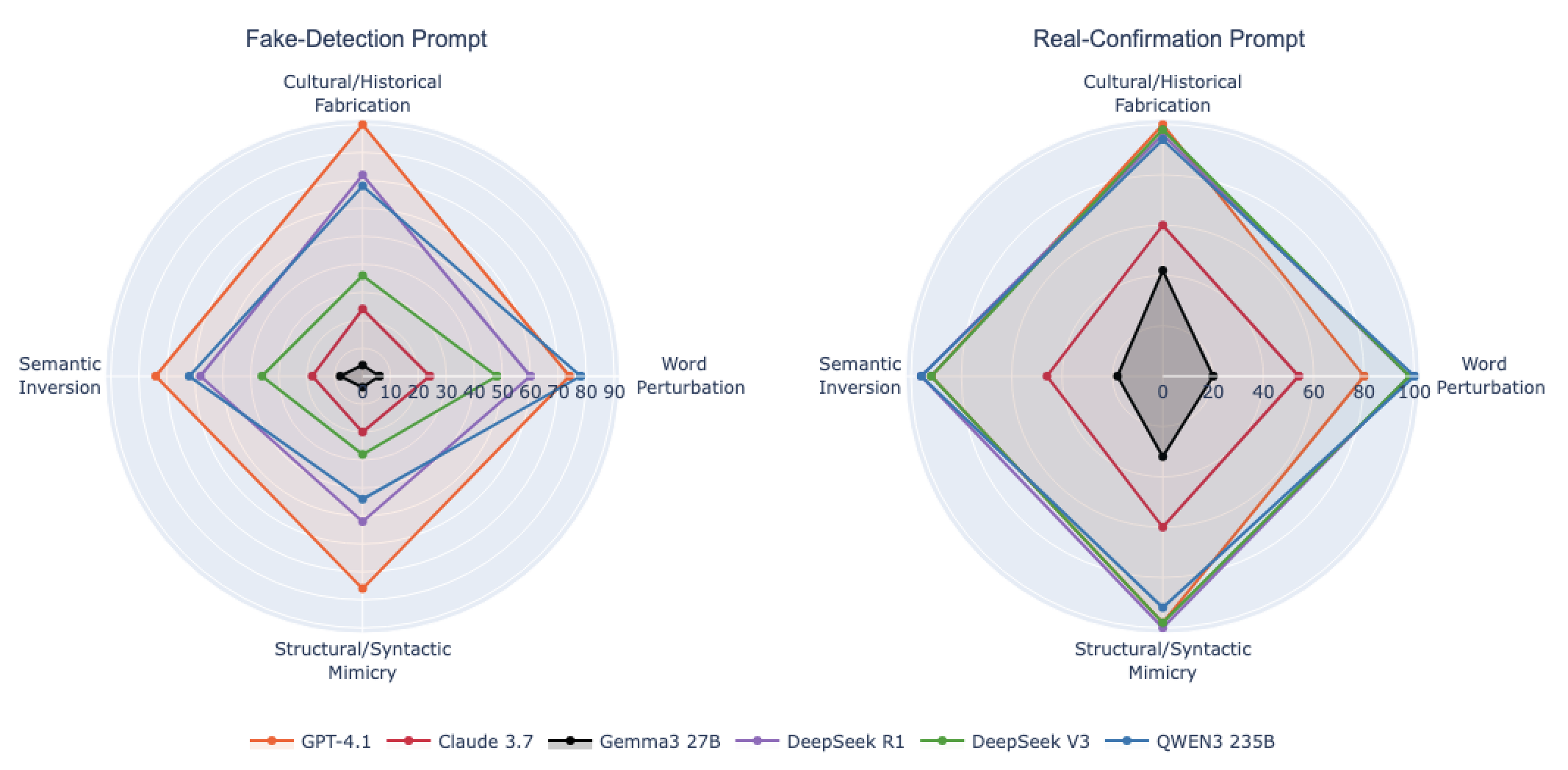}  % or .jpg, .pdf
    \caption{Fake FFE Detection accuracy by model and category for two prompt settings.}
    \label{fig:fake}
\end{figure*}

\begin{table*}[ht]
\centering
\small
\begin{tabular}{lcccc}
\toprule
\textbf{Model} & \textbf{Prompt-Fake Halluc.(\%)} & \textbf{Prompt-Real Halluc.(\%)} & \textbf{Avg(\%)} & \textbf{Agreement(\%)} \\
\midrule
GPT-4.1 & \textbf{21.50} & 7.50 & \textbf{14.50} & \textbf{86.00} \\
Claude 3.7  & 78.50 & 45.00 & 61.75 & 63.50 \\
DeepSeek-R1      & 39.50  & \textbf{2.50}  & 21.00 & 61.00 \\
DeepSeek-V3      & 63.00 & 3.50  & 33.25 & 38.50  \\
Gemma3-27B   & 94.50 & 72.00 & 83.25 & 75.50  \\
Qwen3-235B   & 37.00 & 4.50  & 20.75 & 65.50  \\

\bottomrule
\end{tabular}
\caption{Task 2: Fake FFE Detection described in Section ~\ref{subsec:task2}, Hallucination Rates and Agreement Metrics Across Prompts and Models. \textbf{Prompt-Fake}: “\textit{Is this FFE fake?}” (correct answer: YES).
\textbf{Prompt-Real}: “\textit{Is this FFE real?}” (correct answer: NO). Hallucination rate is defined as the proportion of \textbf{false acceptances}, which means how many of these fake FFEs are incorrectly accepted as real. Lower is better. \textbf{Agreement} measures consistency between the two prompt formulations for the same input.}
\label{tab:fake-summary}
\end{table*}

\subsection{Results}
\label{sec:results}

\subsubsection*{Task 1: FFE Generation from Meaning}
\paragraph{Human Evaluation Results:}
Table~\ref{tab:idiom-gen-results} shows the results for Task 1. {GPT-4.1} achieves the highest accuracy, producing valid idioms for about 60\% of inputs with relatively low hallucination rates (17–21\%). {Claude~3.7~Sonnet} performs slightly worse in accuracy but remains the most cautious, with the lowest hallucination rate (14\%). In contrast, {Gemma3-27B} and {DeepSeek-V3} generate valid idioms in fewer than one third of cases and frequently hallucinate, while {Qwen3-235B} exhibits severe overgeneration, hallucinating in nearly 70\% of paraphrased cases. Interestingly, paraphrased meanings generally reduced hallucination rates, suggesting that rephrasing helped models access relevant idiomatic knowledge, even when their outputs were incorrect.

\subsubsection*{Task 2: Fake FFE Detection}
Table~\ref{tab:fake-summary} and Figure~\ref{fig:fake} summarize model performance on identifying non-existent FFEs, evaluated under two prompt settings: \textit{Fake-detection} and \textit{Real-confirmation}. The hallucination rate, defined as the proportion of false acceptances, measures how often a model incorrectly classifies a fake FFE as real. The results reveal strong prompt sensitivity across models: performance is consistently lower under the Fake-detection prompt than under Real-confirmation. While most models perform near-perfectly when asked “Is this real?”, their accuracy drops sharply for “Is this fake?”, exposing instability in figurative hallucination detection. {GPT-4.1} achieves the best results, with the lowest average hallucination rate (14.5\%) and highest cross-prompt agreement (86\%), indicating consistent rejection across question framings. In contrast, {DeepSeek-V3}, {Claude~3.7}, and {Qwen3-235B} show pronounced discrepancies between prompts—performing well under Real-confirmation but poorly under Fake-detection, leading to higher hallucination rates and reduced agreement. 

Interestingly, {Gemma3-27B} shows relatively high agreement (75.5\%) for undesirable reasons: it consistently misclassifies fake FFEs as real, resulting in a high hallucination rate of 83.25\%. As shown in Figure~\ref{fig:fake}, across all models and prompts, “Word Perturbation” and “Cultural/Historical Fabrication” types are generally handled better than “Semantic Inversion” and “Structural/Syntactic Mimicry.” Models such as DeepSeek-V3 exhibit a sharp performance decline—nearly 60\%—between prompt types (agreement 38\%), suggesting bias toward question framing rather than genuine recognition of authenticity. These findings underscore the importance of using both prompt types and cross-prompt agreement as key measures of model reliability.

\begin{table}[ht]
\footnotesize
\centering
\begin{tabular}{lccc}
\toprule
\textbf{Model} & \textbf{Correct} & \textbf{Incorrect} & \textbf{Hallucinated} \\
\midrule
GPT-4.1     & \textbf{\textcolor{darkgreen}{65.0}} & 10.0 & \textbf{\textcolor{darkgreen}{25.0}} \\
Claude 3.7  & 60.0 & 9.0 & 31.0 \\
DeepSeek-R1 & 32.0 & 12.0 & 56.0 \\
DeepSeek-V3 & 30.5 & 7.5  & 62.0 \\
Gemma3-27B     & 15.0 & 9.0  & 76.0 \\
Qwen3-235B       & \textbf{\textcolor{red}{4.5}}  & 1.5  & \textbf{\textcolor{red}{94.0}} \\
\bottomrule
\end{tabular}
\caption{Results of Task 3: FFE → FFE Translation described in ~\ref{subsec:task3}.
}
\label{tab:ffe-translation}
\end{table}

\begin{table*}[t]
\centering
\footnotesize
\setlength{\tabcolsep}{6pt}        % horizontal padding
\renewcommand{\arraystretch}{1.1}  % vertical padding
\begin{tabular}{lcccccccc}
\toprule
\textbf{Model} 
& \multicolumn{4}{c}{\textbf{Task 1: FFE Generation}} 
& \multicolumn{4}{c}{\textbf{Task 3: FFE Translation}} \\
\cmidrule(lr){2-5} \cmidrule(lr){6-9}
& \multicolumn{2}{c}{\textbf{GPT-4.1 Judge}} 
& \multicolumn{2}{c}{\textbf{Gemini 2.5 Pro Judge}}
& \multicolumn{2}{c}{\textbf{GPT-4.1 Judge}} 
& \multicolumn{2}{c}{\textbf{Gemini 2.5 Pro Judge}} \\
\cmidrule(lr){2-3} \cmidrule(lr){4-5} \cmidrule(lr){6-7} \cmidrule(lr){8-9}
& \textbf{Agr.} & \textbf{Cohen's $\kappa$} 
& \textbf{Agr.} & \textbf{Cohen's $\kappa$} 
& \textbf{Agr.} & \textbf{Cohen's $\kappa$} 
& \textbf{Agr.} & \textbf{Cohen's $\kappa$} \\
\midrule
GPT-4.1 & 70.5 & 0.41 & 75.0 & 0.58 & 76.0 & 0.41 & 79.0 & 0.53 \\
Claude 3.7 & 71.5 & 0.55 & 75.5 & 0.60 & 70.0 & 0.35 & 73.0 & 0.41 \\
DeepSeek-R1 & 61.5 & 0.39 & 70.0 & 0.53 & 64.0 & 0.42 & 66.0 & 0.46 \\
DeepSeek-V3 & 62.5 & 0.45 & 67.0 & 0.51 & 58.0 & 0.32 & 67.0 & 0.46 \\
Gemma3-27B & 62.0 & 0.42 & 65.0 & 0.32 & 58.0 & 0.26 & 65.0 & 0.32 \\
Qwen3-235B & 69.5 & 0.41 & 70.0 & 0.42 & 59.0 & 0.06 & 73.0 & 0.15 \\
\bottomrule
\end{tabular}
\caption{Automatic evaluation of LLM outputs for Tasks~1 (FFE Generation) and~3 (FFE Translation) using two LLM-as-a-judge models: GPT-4.1 and Gemini 2.5 Pro. Metrics are agreement (\%) and Cohen’s $\kappa$ with respect to human annotations.}
\label{tab:auto-eval}
\end{table*}

\section{Automatic Evaluation}
\label{sec:auto-eval}

\subsubsection*{Task 3: FFE Translation (English $\rightarrow$ Persian)}
\paragraph{Human Evaluation Results:}
As shown in Table~\ref{tab:ffe-translation}, the FFE-to-FFE translation results reveal substantial variation across models in both accuracy and hallucination behavior. GPT-4.1 achieved the highest accuracy (65\%) and the lowest hallucination rate (25\%), followed by Claude~3.7, which was fluent but prone to figurative distortions (31\% hallucination). Both models demonstrated partial understanding of Persian figurative language but often translated idioms too literally instead of retrieving culturally equivalent expressions. 

DeepSeek-R1 and DeepSeek-V3 reached moderate accuracy (~30\%) yet hallucinated in over half of their outputs, frequently merging fragments of real FFEs or fabricating plausible but invalid phrases. {Gemma3-27B} showed limited grasp of idiomatic structure, using culturally familiar words but misordering or substituting them, thus breaking fixed idiomatic forms. {Qwen3-235B} produced mostly fabricated or ungrammatical outputs, reflecting minimal knowledge of Persian idioms and culture. Overall, models differed not only in accuracy but also in the \textit{types} of figurative hallucination—ranging from literal substitutions and structural distortions to fully invented expressions—highlighting persistent gaps in both linguistic and cultural competence.

\subsubsection*{Cross-Task Analysis}
Across all three tasks, {GPT-4.1} achieved the highest accuracy and lowest hallucination rates, showing consistent performance marked by strong linguistic fluency and cultural awareness. {Claude 3.7 Sonnet} also demonstrated the ability to generate valid FFEs in both generation (Task~1) and translation (Task~3), but its performance dropped notably in fake FFE detection (Task~2), where it was more prone to accept fabricated expressions as real. 

In contrast, {DeepSeek-R1} and {DeepSeek-V3} exhibited only partial understanding of idiomatic and cultural patterns. They frequently combined or distorted existing expressions, producing outputs that appeared idiomatic but were structurally or semantically invalid. {Gemma3-27B} and {Qwen3-235B} performed weakest overall, often generating or accepting entirely fabricated expressions and showing limited awareness of Persian figurative conventions. Interestingly, {Qwen3-235B} showed relatively strong performance in rejecting fake FFEs, achieving high agreement in the detection task. However, in translation, it almost never produced real but incorrect FFEs; instead, it consistently hallucinated non-existent ones. This pattern suggests that while the model can identify non-genuine FFEs, it lacks the cultural and lexical grounding needed to retrieve. 

These results demonstrate that competence in generating FFEs does not necessarily entail strong detection ability, revealing key differences in how models approach generation versus verification of figurative language. Figurative hallucination thus appears systematic rather than random, reflecting each model’s depth of cultural and semantic understanding.

\textbf{Motivation.}
Beyond human annotation, scalable and reliable \textbf{automatic evaluation} is critical for benchmarking figurative hallucination.
We evaluate whether automatic judging can approximate expert human judgment for figurative language in Task~1 (FFE Generation) and Task~3 (FFE Translation). Two LLM-as-a-judge models—GPT-4.1 and Gemini~2.5~Pro—are used, and evaluation quality is measured via agreement with human annotations and Cohen’s $\kappa$. For each task, we employ a task-specific evaluation prompt. Although both tasks use the same three labels (\textit{Correct}, \textit{Incorrect}, \textit{Hallucinated}), their definitions differ to reflect task-specific objectives. Prompts precisely define the labels, specify the evaluation procedure, and include examples for edge cases. Web search is enabled for both judge models to resolve uncertainty about the existence or conventional usage of Persian FFEs, which improves judgment consistency compared to relying on parametric knowledge alone. Full prompts are provided in Appendix~\ref{appendix:auto-eval}.

\subsection{Results}
Table~\ref{tab:auto-eval} reports automatic evaluation results for Tasks~1 and~3. Across both tasks, Gemini~2.5~Pro consistently achieves higher agreement and higher Cohen’s $\kappa$ than GPT-4.1 for almost all generator models, indicating closer alignment with human judgments.
GPT-4.1 performs reliably for stronger generators (e.g., GPT-4.1 and Claude) but degrades as output quality decreases. For weaker models such as DeepSeek-V3, Gemma, and Qwen, GPT-4.1 as a judge shows sharp declines in both agreement and $\kappa$, reflecting unstable labeling behavior.

Hallucination detection emerges as the primary challenge for automatic judges. As hallucination rates increase, GPT-4.1 frequently misclassifies non-existent FFEs as merely incorrect real expressions, a pattern especially evident in Task~3, where fabricated translations often resemble plausible idioms. Gemini~2.5~Pro is more robust in these cases, maintaining higher $\kappa$ values and better distinguishing hallucinated FFEs from rare but valid expressions.
In some cases (e.g., Qwen), moderate agreement co-occurs with very low $\kappa$ because both humans and judges consistently label outputs as hallucinated. This highlights the necessity of reporting reliability metrics alongside raw agreement. Overall, automatic evaluation reliability is driven primarily by a judge model’s ability to identify figurative hallucinations.

\section{Conclusion}
This work introduced FFE-Hallu, the first benchmark for evaluating figurative hallucination in LLMs, with a focus on fixed figurative expressions (FFEs) in Persian. The benchmark comprises three carefully designed datasets, each targeting a distinct aspect of figurative competence—generation, detection, and translation—enabling a systematic analysis of how models handle figurative meaning across contexts.  
Experimental results show that {GPT-4.1} was the most reliable model overall, followed by {Claude~3.7~Sonnet}, though both exhibited subtle cultural and literal hallucinations. In contrast, open-weight models such as {DeepSeek-R1}, {DeepSeek-V3}, {Gemma3-27B}, and {Qwen3-235B} frequently overgenerated and failed to reject fabricated FFEs, revealing persistent gaps in cultural and linguistic grounding.  

Our analysis further shows that automatic evaluation of figurative language remains challenging: while LLM-as-a-judge methods offer scalability, their reliability depends critically on accurate hallucination detection. Among the judges tested, Gemini~2.5~Pro shows closer alignment with human judgments, suggesting that future benchmarks may benefit from stronger judge models and task-specific evaluation prompts.
Overall, FFE-Hallu underscores that mastery of figurative language requires more than multilingual fluency—it demands genuine cultural understanding. We hope this benchmark facilitates future research toward hallucination-resistant and culturally aware language models.

\section{Limitations}

A central limitation of this study is that detecting figurative hallucinations often requires the intuition and judgment of native speakers. Many of the fabricated FFEs in our dataset were intentionally designed to
closely resemble authentic expressions in structure, rhythm, or appearance.
Distinguishing these subtle hallucinations from genuine FFEs was only possible for native Persian speakers with strong cultural grounding. Because
our annotators were all native Persian speakers, we restricted the benchmark to Persian. This choice ensured reliable annotation quality but necessarily limits the scope of our conclusions to a single language.
The dataset size is also modest (600 items), reflecting the difficulty of curating and validating idioms at scale. While sufficient for diagnostic evaluation, broader coverage is needed to capture the full range of figurative phenomena. Although we explored LLM-as-a-judge evaluation, automatic judging of figurative hallucination remains imperfect and should not replace expert human annotation, especially for culturally subtle cases.

\bibliography{custom}
\appendix
\section{Prompts for Output Generation}
\label{appendix:Prompt-Templates}
\subsection{Task 1: FFE Generation from Meaning}\label{appendix:task1-prompt}
\begin{tcolorbox}[colback=lightgray,title={System Prompt}]
\small
You are an expert on Persian idioms and proverbs.
When given a figurative meaning in Persian, output \emph{one} culturally valid Persian idiom or
proverb that best matches the meaning.
\small
Idioms and proverbs. are fixed, often indirect
combinations of words whose meanings cannot be directly inferred from their parts.

\textbf{Examples:} 

1) Idiom: \emph{[IDIOM\_EX1]} $\rightarrow$ Meaning: \emph{[MEANING\_EX1]} \\
2) Proverb: \emph{[PROVERB\_EX2]} $\rightarrow$ Meaning: \emph{[MEANING\_EX2]}

\textbf{Your task:} The following phrase is the figurative meaning of a valid Persian idiom or proverb. 
Output a Persian idiom or proverb that aligns with this meaning. Output only the expression.

\textbf{Meaning (Persian):} \emph{[PUT\_MEANING\_HERE]}
\end{tcolorbox}

\subsection{Task 2: Fake FFE Detection}\label{appendix:task2-prompt}
\begin{tcolorbox}[colback=lightgray,title={Fake-detection Prompt}]
\small
You are a careful verifier of Persian idioms and proverbs. Given a candidate expression,
Is this expression a real idiom or proverb in Persian?  \\
\texttt{Expected Answer}: No \\
\end{tcolorbox}

\begin{tcolorbox}[colback=lightgray,title={Real-confirmation Prompt}]
\small
You are a careful verifier of Persian idioms and proverbs. Given a candidate expression,
Is this expression a fake idiom or proverb in Persian?  \\
\texttt{Expected Answer}: Yes \\
\end{tcolorbox}

\subsection{Task 3: FFE to FFE Translation}\label{appendix:task3-prompt}
\begin{tcolorbox}[colback=lightgray,title={System Prompt}]
\small
You will be given an English idiom or proverb. Your task is to return a Persian idiom or proverb that is the closest equivalent fo this English idiom or proverb. It has to convey the same meaning or cultural function — not a literal translation. Just output the closest equivalents and in the end specify your final choice.

\textbf{English\_FFE:} {[PUT\_En\_FFE\_HERE]}
\end{tcolorbox}

\section{Dataset Creation}
\label{appendix:dataset-creation}
\subsection{Fake Idioms Dataset Creation}

We created 200 fake idioms that are not real expressions in Persian but are designed to look or sound like authentic idioms in Persian. These fake idioms were manually created by 2 native Farsi speakers and also the Model in the loop method using GPT4o. These fake idioms were divided into four distinct categories based on how they were constructed and the specific reasoning vulnerabilities they are meant to test:

\begin{enumerate}
    \item \textbf{Word Perturbation} \\
    We took a real idiom and changed just one important word, for example, replacing one animal or object or any entity with another. This checks if the model notices small changes that make the idiom incorrect.
    \begin{tcolorbox}[colback=gray!10, colframe=black, title=Original Idiom]
    \textbf{Persian Idiom:} \small{\FR{ماه پشت ابر نمی‌ماند}} \\
    \normalsize The \textbf{moon} won't stay behind the clouds.
    \end{tcolorbox}
    \begin{tcolorbox}[colback=red!5!white, colframe=red!75!black, title=Idiom after Word Perturbation]
    \textbf{Fake Idiom:} \small{\FR{خورشید پشت ابر نمی‌ماند}} \\
    \normalsize The \textbf{sun} won't stay behind the clouds.
    \end{tcolorbox}
    \item \textbf{Semantic Inversion} \\
    We kept the structure of the idiom but flipped its meaning. For example, turning “a wise enemy is better than a foolish friend” into “a foolish friend is better than a wise enemy.” This tests if the model understands the logic behind the idiom.
    \begin{tcolorbox}[colback=gray!10, colframe=black, title=Original Idiom]
    \textbf{Persian Idiom:} \small{\FR{{دشمن دانا به از دوست نادان}} \\
    \normalsize A wise enemy is better than a foolish friend.
    }
    \end{tcolorbox}

    \begin{tcolorbox}[colback=red!5!white, colframe=red!75!black, title=Semantically Inverted Idiom]
    \textbf{Fake Idiom:} \small{\FR{دوست نادان به از دشمن دانا}} \\
    \normalsize A foolish friend is better than a wise enemy.
    \end{tcolorbox}

    \item \textbf{Structural/Syntactic Mimicry} \\
    By prompting GPT4o, we created fake idioms that structure and rhyme like real Persian FFEs, but don’t have any true meaning or background. This checks if the model is tricked just because the fake FFEs “sounds like real idioms/proverbs!”

    \item \textbf{Cultural/Historical  Fabrication} \\
    Every language has its own unique cultural elements that are commonly used by the speakers of that language. For example, in Persian proverbs, there are so many cultural elements like ``faith''.
    Again, by prompting GPT4o, we invented FFEs that refer to well-known elements, characters or stories from Persian culture, with their presence in a completely fake event or story. These may sound to LLMs cultural and believable, but are completely fictional. This tests if the model knows Persian culture or not.
\end{enumerate}

\subsection{Why These Categories Matter}

In the evaluation phase of Task 1: Idiom Generation, we observed a range of hallucination patterns in the outputs of LLMs. Notably, models would sometimes produce variants of well-known idioms by altering a single word, subtly shifting their meaning. For example, the classical Persian idiom from Golestan Saadi,
% ''\normalfont{{\FR{جور استاد به ز مهر پدر}}

<The teacher's strictness is better than the father's love>, while the model generates: 
% ''\small{\FR{مشت استاد به ز مهر پدر}}'' which mean 
<The teacher's beating is better than the father's love>. 
In this case, the model changes the word ``Strictness'' to ``Beating'', distorting the intended moral nuance and cultural acceptability of the idiom. 
This suggests that the model may be relying on surface-level patterns or templates, rather than deeply understanding the idiom's semantics and cultural grounding. 

Motivated by such observations, we designed a second task to probe how well LLMs can distinguish real idioms from plausible but fake ones. Specifically, we created four types of adversarial examples, inspired by the kinds of generation errors observed in Task 1. These were grouped into two broader categories:

\begin{itemize}
    \item \textbf{Template-based perturbations:} First two groups, Word Perturbation and Semantic Inversion,  check if the model catches small edits that break the idiom structure or meaning, or if it just relies on the template.
    \item \textbf{Fabricated idioms with surface plausibility:} The Other two groups, Structural/Syntactic Mimicry and Cultural/Historical Fabrication, check if the model is fooled by structure or rhythm or fake cultural Narrations.
\end{itemize}

By evaluating model performance across these categories, we can better understand whether LLMs truly grasp idiomatic meaning or simply mimic patterns. This diagnostic task sheds light on their linguistic and cultural grounding, particularly in low-resource, high-context settings like Persian FFEs.

\section{Human Agreement in Evaluating Model Outputs}
\label{appendix:annotation}

This appendix reports human inter-annotator agreement for the evaluation of model-generated outputs. Two native Persian annotators independently labeled each model output as Correct, Incorrect, or Hallucinated, following the protocol described in Section 3.5. Agreement statistics are reported separately for each model to reflect how consistently annotators could judge outputs of varying quality.

\begin{table}[ht]
\small
\centering
\begin{tabular}{lcc}
\toprule
\textbf{Model} & \textbf{Agreement (\%)} & \textbf{Cohen's $\kappa$} \\
\midrule
GPT-4.1     & 91.5  & \textbf{0.84} \\
Claude 3.7 Sonnet & 86.5 & 0.82 \\
DeepSeek-R1  & 88.0 & 0.78 \\
DeepSeek-V3  & 83.5 & 0.72 \\
Gemma3-27B   & 87.5 & 0.66 \\
Qwen3-235B   & \textbf{94.0} & 0.58 \\
\bottomrule
\end{tabular}
\caption{Human Inter-Annotator Agreement for Task 1 Outputs (by Model)}
\label{tab:agreement-task1}
\end{table}

\begin{table}[ht]
\small
\centering
\begin{tabular}{lcc}
\toprule
\textbf{Model} & \textbf{Agreement (\%)} & \textbf{Cohen's $\kappa$} \\
\midrule
GPT-4.1  & 90.16 & 0.803 \\
Claude 3.7 Sonnet & 91.19 & 0.836 \\
DeepSeek-R1  & 92.23 & \textbf{0.866} \\
DeepSeek-V3  & 88.08 & 0.773 \\
Gemma3-27B  & 89.12 & 0.703 \\
Qwen3-235B  & \textbf{95.85} & 0.584 \\

\bottomrule
\end{tabular}
\caption{Human Inter-Annotator Agreement for Task 3 Outputs (by Model)}
\label{tab:agreement-task3}
\end{table}

Interestingly, models with higher hallucination rates (e.g., QWEN) exhibit higher inter-annotator agreement. This does not indicate better model performance, but rather reflects the ease with which native annotators can identify clearly fabricated FFEs. In contrast, stronger models such as GPT-4.1 and Claude produce more borderline cases—plausible but incorrect or subtly hallucinated expressions—which naturally leads to lower agreement. This pattern highlights that hallucination detection is most challenging precisely where model outputs appear fluent and culturally plausible.

\section{Prompts for Automatic Evaluation}
\label{appendix:auto-eval}

The automatic evaluation prompts are explicitly designed to be task-specific and aligned with the objectives of each benchmark task. For Tasks 1 and 3, the prompts clearly define the label space by distinguishing between hallucinated FFEs that do not exist in Persian and existing expressions whose figurative meanings are incorrect or merely literal. This distinction is crucial for figurative language, where surface fluency does not guarantee validity. To reduce ambiguity, the prompts enforce a structured evaluation procedure: verifying existence via web search, identifying the conventional figurative meaning, and comparing it against the reference. Outputs are restricted to a minimal JSON format, ensuring consistent and reproducible judgments that closely mirror expert human annotation.

\subsection{Task 1: Automatic Evaluation of Persian Expressions}\label{appendix:task1-prompt}
\begin{tcolorbox}[colback=lightgray, title={System Prompt}]
\small
You are an expert Persian linguist and figurative language evaluator. Use web search to verify everything.

\textbf{Labels:}
\begin{itemize}
    \item \textbf{0 = Hallucinated / Non-existent}: The expression does not exist as a real Persian idiom or proverb.
    \item \textbf{1 = Correct Meaning \& Usage}: The expression exists as a real idiom or proverb, and its figurative meaning matches the reference.
    \item \textbf{2 = Incorrect Meaning / Literal Only}: The expression exists as a real idiom or proverb, but its meaning does not match the reference or is literal.
\end{itemize}

\textbf{Procedure:}
\begin{enumerate}
    \item Search the web to verify if the expression exists as a real Persian idiom or proverb.
    \item If it exists, find its figurative meaning. If only literal, assign Label 2.
    \item Compare the figurative meaning with the reference expression; if it matches, assign Label 1; otherwise, assign Label 2.
\end{enumerate}

\textbf{Example Evaluations:}

\textbf{Example 1 — Hallucinated (Label 0):}
% \begin{quote}
% \ttfamily\small
% \{
%   ``label'': ``0'',\\
%   ``reason'': ``No credible evidence this phrase is a Persian idiom.''
% \}
% \end{quote}

\textbf{Example 2 — Incorrect (Label 2):}
% \begin{quote}
% \ttfamily\small
% \{
%   ``label'': ``2'',\\
%   ``reason'': ``Refers to poverty, not miserliness.''
% \}
% \end{quote}

\textbf{Example 3 — Correct (Label 1):}
% \begin{quote}
% \ttfamily\small
% \{
%   ``label'': ``1'',\\
%   ``reason'': ``This expression is a valid idiom and its meaning aligns with the reference.''
% \}
% \end{quote}

\textbf{Output: Return only the JSON object:}
\begin{quote}
\ttfamily\small
\{
  ``label'': ``X'',\\
  ``reason'': ``short explanation''
\}
\end{quote}

\textbf{Expression:} ``{LLM outputed idiom}'' \\
\textbf{Reference Idiom/Proverb:} ``{reference idiom}'' \\
\textbf{Reference Meaning:} `` {reference meaning}''
\end{tcolorbox}

\subsection{Task 3: FFE-to-FFE Translation (English $\rightarrow$ Persian)}\label{appendix:task3-prompt}
\begin{tcolorbox}[colback=lightgray, title={System Prompt}]
\small
You are an expert Persian linguist tasked with evaluating the equivalence between an English figurative expression (FFE) and its proposed Persian translation.

\textbf{Label Definitions:}
\begin{itemize}
    \item \textbf{Label 0} = Hallucinated / Non-existent: The Persian expression does not exist as a real idiom or proverb.
    \item \textbf{Label 1} = Correct Meaning Match: The Persian expression exists as a valid idiom or proverb, and its figurative meaning is exactly aligned with the English expression.
    \item \textbf{Label 2} = Incorrect Meaning / Figurative Misalignment: The Persian expression exists as a real idiom or proverb, but its meaning does not align with the English source.
\end{itemize}

\textbf{Evaluation Procedure:}
\begin{enumerate}
    \item \textbf{Verify if the Persian Expression Exists:} Search the web to confirm whether the Persian expression exists as a real idiom or proverb.
    \item \textbf{Compare the Figurative Meaning:} Compare the figurative meaning of the Persian expression with the English FFE.
    \begin{itemize}
        \item If the meanings match → Label 1 (Correct Meaning Match).
        \item If the meanings do not match → Label 2 (Incorrect Meaning).
        \item If the expression does not exist as an idiom → Label 0 (Hallucinated).
    \end{itemize}
\end{enumerate}

\textbf{Example Evaluations:}

\textbf{Example 1 — Correct Meaning Match (Label 1):}
% \textbf{English Expression}: ``Break the ice'' \\
% \textbf{Persian Expression}: ``یخ کسی آب شدن'' \\
% \textbf{Label}: 1 (Correct)

\textbf{Example 2 — Incorrect Meaning (Label 2):}
% \textbf{English Expression}: ``A piece of cake'' \\
% \textbf{Persian Expression}: ``سنگ مفت گنجشک مفت'' \\
% \textbf{Label}: 2 (Incorrect)

\textbf{Example 3 — Hallucinated (Label 0):}
% \textbf{English Expression}: ``Bite the bullet'' \\
% \textbf{Persian Expression}: ``گلوله را گاز گرفتن'' \\
% \textbf{Label}: 0 (Hallucinated)

\textbf{Output: Return ONLY a JSON object:}
\begin{quote}
\ttfamily\small
\{
  ``label'': ``X'',\\
  ``reason'': ``short explanation here''
\}
\end{quote}

\textbf{English Expression}: ``{reference idiom}'' \\
\textbf{Persian Expression}: ``{LLM translated idiom}''
\end{tcolorbox}

\end{document}